\documentclass[conference]{IEEEtran}
\IEEEoverridecommandlockouts
\usepackage{cite}
\usepackage{amsmath,amssymb,amsfonts}
\usepackage{algorithmic}
\usepackage{graphicx}
\usepackage{textcomp}
\usepackage{xcolor}
\usepackage{booktabs}
\usepackage{makecell}
\usepackage{tikz}
\def\BibTeX{{\rm B\kern-.05em{\sc i\kern-.025em b}\kern-.08em
    T\kern-.1667em\lower.7ex\hbox{E}\kern-.125emX}}

\newcommand\copyrighttext{%
  \footnotesize \textcopyright 2023 IEEE.  Personal use of this material is permitted.  Permission from IEEE must be obtained for all other uses, in any current or future media, including reprinting/republishing this material for advertising or promotional purposes, creating new collective works, for resale or redistribution to servers or lists, or reuse of any copyrighted component of this work in other works.}
\newcommand\copyrightnotice{%
\begin{tikzpicture}[remember picture,overlay]
\node[anchor=south,yshift=10pt] at (current page.south) {\fbox{\parbox{\dimexpr\textwidth-\fboxsep-\fboxrule\relax}{\copyrighttext}}};
\end{tikzpicture}%
}

\begin{document}

\title{Relationship between Model Compression and Adversarial Robustness: A Review of Current Evidence}

\author{\IEEEauthorblockN{Svetlana Pavlitska$^{1,2}$, Hannes Grolig$^{2}$, J.~Marius Zöllner$^{1,2}$}
\IEEEauthorblockA{\textit{$^{1}$ FZI Research Center for Information Technology} \\
 \textit{$^{2}$ Karlsruhe Institute of Technology (KIT)}\\
Karlsruhe, Germany \\
pavlitska@fzi.de}
}

% \author{\IEEEauthorblockN{1\textsuperscript{st} Given Name Surname}
% \IEEEauthorblockA{\textit{dept. name of organization (of Aff.)} \\
% \textit{name of organization (of Aff.)}\\
% City, Country \\
% email address or ORCID}
% \and
% \IEEEauthorblockN{2\textsuperscript{nd} Given Name Surname}
% \IEEEauthorblockA{\textit{dept. name of organization (of Aff.)} \\
% \textit{name of organization (of Aff.)}\\
% City, Country \\
% email address or ORCID}
% \and
% \IEEEauthorblockN{3\textsuperscript{rd} Given Name Surname}
% \IEEEauthorblockA{\textit{dept. name of organization (of Aff.)} \\
% \textit{name of organization (of Aff.)}\\
% City, Country \\
% email address or ORCID}
% \and
% \IEEEauthorblockN{4\textsuperscript{th} Given Name Surname}
% \IEEEauthorblockA{\textit{dept. name of organization (of Aff.)} \\
% \textit{name of organization (of Aff.)}\\
% City, Country \\
% email address or ORCID}
% \and
% \IEEEauthorblockN{5\textsuperscript{th} Given Name Surname}
% \IEEEauthorblockA{\textit{dept. name of organization (of Aff.)} \\
% \textit{name of organization (of Aff.)}\\
% City, Country \\
% email address or ORCID}
% \and
% \IEEEauthorblockN{6\textsuperscript{th} Given Name Surname}
% \IEEEauthorblockA{\textit{dept. name of organization (of Aff.)} \\
% \textit{name of organization (of Aff.)}\\
% City, Country \\
% email address or ORCID}
% }

\maketitle
\copyrightnotice
\thispagestyle{empty}
\pagestyle{empty}

\begin{abstract}
Increasing the model capacity is a known approach to enhance the adversarial robustness of deep learning networks. On the other hand, various model compression techniques, including pruning and quantization, can reduce the size of the network while preserving its accuracy. Several recent studies have addressed the relationship between model compression and adversarial robustness, while some experiments have reported contradictory results. This work summarizes available evidence and discusses possible explanations for the observed effects.
\end{abstract}


%%%%%%%%% BODY TEXT
\section{Introduction and Related Work}
\label{sec:intro}

Goodfellow et al.~\cite{explaining2014} and Szegedy et al.\cite{szegedy2013intriguing} first brought up the risk of adversarial attacks, small perturbations (often imperceptible by humans) that are carefully crafted and added to the input of state-of-the-art (SOTA) deep neural networks (DNNs). Without specific DNN training or mitigation measures, these attacks lead to high-confidence wrong outputs of SOTA DNNs and convolutional neural networks (CNNs). %There has been a great interest in the scientific community to defend against numerous adversarial attacks. The more DNNs are deployed, the higher the incentives for an adversary to fool them for malicious reasons. 
This inherent vulnerability of DNNs poses an especially high risk when applying them in autonomous driving, facial recognition, or medical domains.

Adversarial defenses attempt to robustify neural networks artificially, but robustly solving a task fundamentally increases its difficulty. %For example, classifying images is much more difficult if an adversarial attacker can search for weaknesses and uncertainties in the decision process. 
%It comes thus as no surprise that more capable networks while requiring significantly more compute resources, can generally solve tasks more robustly.
However, simply scaling model sizes is not always an option and is quickly restricted by technical and financial factors. Model compression approaches such as quantization and pruning can significantly reduce model size while preserving comparable performance levels. 

The impact of model compression on adversarial robustness has been a focus of several recent studies. However, to the best of our knowledge, no analysis of the existing publications to summarize the state of the art has been performed so far. Our work aims at closing this research gap. We have reviewed existing works that either explored the effect of model compression methods on the adversarial vulnerability of the networks or tried to combine both goals in a single training algorithm. We group the existing evidence from the experiments and make conclusions based on these.

\section{Related Work}

\subsection{Adversarial Training}
%Research on adversarial robustness has been an arms race of algorithms to defend against adversarial examples, followed by more advanced algorithms to craft adversarial examples rendering previous defenses useless. %Adversarial training (AT) has the disadvantage of being more expensive in terms of time complexity than standard training. This is why the search for ways to robustify models against adversarial attacks has brought up other training branches like regularization methods (defensive distillation \cite{distillation1, distillation2}, label smoothing, feature squeezing \cite{featuresqueezing1,featuresqueezing2}) and adversarial detection algorithms \cite{detection}. %Other strategies like Generative Adversarial Networks (GANs), logit-squeezing, or Jacobian regularization have not yet been applied to large-scale problems such as ImageNet, according to \cite{shafahi_adversarial_2019}. %The so-called certified or provable defenses could also not be demonstrated to be computationally efficient or satisfiably performant for large-scale networks and high perturbation levels. 
Adversarial training (AT) remains among the most successful defenses against adversarial examples \cite{shafahi_adversarial_2019, pang_bag_2021, maini_adversarial, schott2018towards, athalye_synthesizing_2018}. Salman et al. showed that adversarially trained ImageNet~\cite{russakovsky2015imagenet}-classifiers show better transferability \cite{SalmanIEKM20}, which is consistent with the hypothesis that adversarially trained robust networks provide better feature representations. Gong et al. showed that AT can improve image recognition models by preventing overfitting \cite{aes_xie}. Andriushchenko et al. \cite{Andriushchenko_understanding_20} stated that performing AT efficiently is important because it is the crucial algorithm for robust deep learning. The idea is intuitive: DNNs are trained by handing them data and correct labels to learn their decision boundaries. In AT, adversarial examples and their correct labels are precautiously augmented into the training process to train a more robust model. Madry et al. proposed the prime baseline for AT with a Projected Gradient Descent (PGD) attack \cite{madry2017towards}, which was later improved by \cite{rice_overfitting_2020} using early-stopping.% Due to the high computational cost of crafting multi-step PGD AEs for large datasets like ImageNet, these AT algorithms were only within reach for research groups with enormous GPU Power. Consequently, the field of fast/efficient AT has emerged with the goal to speed up AT by reducing the complexity of generating AEs. It is undiscussed that AT still has to improve for DNNs to be rolled out in critical infrastructure and autonomous driving scenarios. On the small MNIST dataset, Madry et al. \cite{madry2017towards} train a model with 90\% accuracy under attack. But for more complex datasets like CIFAR-10 (best accuracy under attack around 57\%) or even ImageNet (best accuracy under attack around 49\%), SOTA AT approaches do not get past 60\%, which is still unsatisfactory.

% \section{Related Work}
% \subsection{Adversarial Attacks}
% First described in \cite{szegedy2013intriguing}. 
% In general, most attacks are described in \cite{tramer2017ensemble}
% "Adversarial attacks typically minimize an lp norm (e.g., l2, linf, l1 and l0) of the required perturbation under certain (box) constraints." \cite{guo2018sparse}
% \\
% \textbf{White-Box, Black-box and Grey-Box Attacks}
% TBD
% \\
% \textbf{Fast Gradient Sign Method (FGSM)}\\
% \cite{explaining2014} + IFGSM + rFGSM in \cite{tramer2017ensemble}
% \\
% \textbf{Sign-based Projected Gradient Descent (PGD)}\\
% \cite{madry2017towards}
% \\
% \textbf{Carlini and Wagner Attack (C\&W Attack)}\\
% \cite{carlini2017towards} describe all attacks ($L_{1}$, $L_{2}$ and $L_{\infty}$)
% \\
% \textbf{WRM Attack}\\
% \cite{sinha2017certifying}
% \\
% \textbf{DeepFool} \\
% \cite{deepfool2016}
% \\
% \textbf{Transfer Attacks}\\
% \cite{ye2019adversarial}
% \\
% \textbf{Random Perturbation (Random)}\\
% \cite{lin2019defensive} + Random attack from \cite{hu2020triple} in Appendix D.
% \\
% \textbf{Semantic-Preserving Transformations}\\
% Literature from \cite{jordao2021effect}
% \\
% \textbf{Simple Occlusions}\\
% Literature from \cite{jordao2021effect}
% \\
% \textbf{Papernot black-box attack} \\
% \cite{papernot2016limitations}
% \\
% \subsection{Adversarial Defenses}
% \textbf{Adversarial Training (AT)}\\
% Adversarial Training from \cite{explaining2014} or \cite{ye2019adversarial} 
% \\ 
% \textbf{Other Known Defense Methods} \\
% See \cite{jordao2021effect} for other defenses

\subsection{Model Compression}

DNN and CNN architectures have become increasingly deep and complex and can require millions of parameters, which leads to slow inference. Many techniques have been developed to speed up inference, including quantization and pruning. %Quantization reduces the precision of network parameters. Pruning, on the other hand, reduces the number of parameters and thus reduces the model size and, depending on the pruning approach, also the calculation speed.

Pruning aims at selecting insignificant parameters that can be removed to make the model smaller while maintaining high prediction accuracy. The simplest approach, magnitude-based pruning, removes weights below a specified threshold value. Instead of pruning individual weights, it is also possible to prune at a higher level of granularity by removing entire feature maps or filters in a CNN. Filters can be removed using data-independent pruning methods based on properties such as their L1 norm \cite{li2016pruning}. Correct pruning can help to speed up the inference without impacting accuracy~\cite{han2016deeop}. Quantization is another method that reduces the precision of the model parameters, e.g., from 32-bit floating point to 8-bit integers. It can be performed on scalars or vectors as demonstrated in \cite{stock2020and}, where the reconstruction error of the activations rather than the weights is minimized. %Although quantization reduces the size of parameters by a factor of 4X, the inference is generally accelerated by a factor of 2-3X, as shown by Google's TensorFlow-Lite \cite{TensorFlowLite} and Nvidia's Tensor RT \cite{Migacz2017}. However, quantization and dequantization introduce a computational overhead. %Knowledge distillation \cite{DBLP:journals/corr/HintonVD15} is another approach where a complex model transfers its knowledge to a smaller model. 

\section{Relationship between Quantization and Robustness}

Quantization has so far been a focus of only a few works exploring adversarial robustness (see Table~\ref{tab:overview-quant}). Our search has revealed a total of four papers~\cite{galloway2017,rakin2018defend,wijayanto2019towards,lin2019defensive}, all of which consider both white-box and black-box attacks, while PGD~\cite{madry2017towards} is a method used in all works. %Furthermore, two more works explored quantization combined with pruning~\cite{gui2019model,wijayanto2019towards}.

One of the first works regarding quantization and adversarial robustness is from Galloway et al. \cite{galloway2017}. 
The authors focused on binarized neural networks where both weights and activations in the hidden layers are  quantized to $\pm1$. Randomized quantization was used. 
They compared full-precision networks to their respective binarized network. % and a respective binarized network with a learned scalar on MNIST and CIFAR-10 against FGSM, PGD, and C\&W $L_{2}$ white-box attacks and the black-box attack from Papernot. They also investigated the effect of AT on binarized networks. 
It was observed that AT is a balancing act with binary models, whereas scaled binary models can benefit from AT. 
Overall, they concluded that binarized networks can slightly improve the robustness against certain attacks. In terms of efficiency, they observed an advantage of the binarized networks over their full-precision equivalents. 

In \cite{rakin2018defend}, Rakin et al. proposed a novel approach where activations are quantized to increase the adversarial robustness of DNNs. 
The approach integrates the quantized activation functions into AT. They proposed a fixed as well as a dynamic activation quantization method. 
For experiments, adversarially trained baseline networks %with LeNet on MNIST and ResNet-18 on CIFAR-10 
were used. 
Then, the authors trained LeNet~\cite{lecun1998gradient} and ResNet-18~\cite{he2016deep} with the fixed and dynamic quantization techniques.
 The models were quantized with different quantization levels (1-, 2- and 3-bit activation).
 The robustness of the fixed and dynamic quantized networks against various attacks (PGD~\cite{madry2017towards}, FGSM~\cite{explaining2014}, Carlini and Wagner (C\&W) attack~\cite{carlini2017towards})  was compared with the robustness of the baseline networks. The authors concluded that fixed and dynamic quantization can increase the robustness.

A further work by Wijayanto et al.~\cite{wijayanto2019towards} proposed an adversarial-aware compression framework for DNNs.
This framework combines pruning, quantization, and encoding. In their experiments, the approach is compared to pruned and quantized networks. It was observed that quantization can improve robustness. 

Another novel quantization method is proposed by Lin et al.~\cite{lin2019defensive}, where an empirical study regarding quantization and robustness was conducted. The authors quantized the activations and compared the naive quantized models to their respective full-precision models. 
They observed that the conventional quantization method is not robust and that input image quantization applied to hidden layers worsens the robustness. 
The proposed defensive quantization approach achieved higher robustness than their full-precision counterparts and improved the accuracy without adversarial attack.

\begin{table*} 
  \centering
  \resizebox{1.0\linewidth}{!}{%
 % {\def\arraystretch{1.5}\tabcolsep=5pt
  \begin{tabular}{|l|c | c|c|c|c|c|c|c|c|c|} 
  \hline
   \textbf{Author} & \textbf{Year} & \textbf{Ref} & \textbf{Architectures} & \textbf{Dataset}  &  \textbf{Baseline} & \textbf{Quantization} &   \makecell{\textbf{Attack}} &   \textbf{Attack Method} & \textbf{AT} \\\hline \hline

   Galloway et al. & 2017 &\cite{galloway2017} & \makecell{Small CNN,\\Wide
ResNet-28-4~\cite{zagoruyko2016wideresnet}} & \makecell{MNIST~\cite{lecun1998gradient} \\ CIFAR-10~\cite{krizhevsky2009learning}} & \makecell{Full-precision\\ networks} & \makecell{Binarization}  & \makecell{White-box \\ Black-Box} & \makecell{FGSM~\cite{explaining2014}, PGD~\cite{madry2017towards}, C\&W $L_{2}$~\cite{carlini2017towards}, \\ Papernot's attack~\cite{papernot2016limitations}} & \makecell{ $\times$}  \\ \hline

   Rakin et al. & 2018 &\cite{rakin2018defend} & \makecell{LeNet~\cite{lecun1998gradient},\\ ResNet-18~\cite{he2016deep}} & \makecell{MNIST~\cite{lecun1998gradient} \\ CIFAR-10~\cite{krizhevsky2009learning}} & \makecell{Full-precision\\ networks with AT\\ (PGD~\cite{madry2017towards})} & \makecell{Quantization of\\activation functions }  & \makecell{White-box,\\Black-box} & \makecell{FGSM~\cite{explaining2014}, PGD~\cite{madry2017towards}, C\&W $L_{2}$~\cite{carlini2017towards}\\ Zeroth Order Optimization~\cite{chen2017zoo},\\Substitute model } & $\checkmark$\\ \hline

   Wijayanto et al. & 2019 &\cite{wijayanto2019towards} & \makecell{Inception-v3\\AlexNet\\MobileNet-v1~\cite{howard2018mobilenets}} & ImageNet~\cite{russakovsky2015imagenet} & \makecell{Models compressed via\\Deep compression~\cite{han2016deeop}\\and incremental network\\quantization (INQ)~\cite{zhou2017incremental},\\ compact and int8 models} & \makecell{Dynamic network\\ surgery~\cite{guo2016dynamic} with  INQ\\ and DEFLATE\\ compression during AT}  & \makecell{White-box \\ Gray-Box} & \makecell{FGSM~\cite{explaining2014}, BIM~\cite{kurakin2017adversarial},\\ Transfer attacks} & $\checkmark$  \\ \hline

    Lin et al. & 2019 &\cite{lin2019defensive} & \makecell{VGG-16~\cite{simonyan2014very},\\
ResNet-28-10~\cite{zagoruyko2016wideresnet},\\Wide ResNet-16-4~\cite{zagoruyko2016wideresnet}} & \makecell{CIFAR-10~\cite{krizhevsky2009learning}\\ SVHN~\cite{netzer2011reading}} & \makecell{Full-precision models\\with AT and feature\\ squeezing~\cite{featuresqueezing1}} & \makecell{Defensive quantization\\with Lipschitz\\ regularization}  & \makecell{White-box} & \makecell{FGSM, R-FGSM~\cite{explaining2014}\\BIM~\cite{kurakin2017adversarial}, PGD~\cite{madry2017towards}} & \makecell{ $\checkmark$}  \\ \hline

    Gorsline et al. & 2019 &\cite{gorsline2021adversarial} & \makecell{MLP with 100\\ hidden neurons} & \makecell{MNIST~\cite{lecun1998gradient}\\2-spiral problem} & \makecell{Full-precision model} & \makecell{Weight quantization}  & \makecell{White-box} & \makecell{FGSM~\cite{explaining2014}} & \makecell{ $\times$}  \\ \hline

    Varghese et al. & 2019 &\cite{varghese2022joint} & \makecell{DeepLabv3+~\cite{chen2018encoder}} & \makecell{Cityscapes~\cite{Cordts2016Cityscapes} ,\\SynPeDS~\cite{stauner2022synpeds}} & \makecell{Non-pruned,\\full-precisions models} & \makecell{Magnitude-based pruning,\\quantization with\\uniform rounding}  & \makecell{N/A} & \makecell{Image Corruptions} & \makecell{ $\times$}  \\ \hline
  \end{tabular}
  }
  \caption{Overview of the publications analyzing the relationship between \textbf{quantization} and adversarial robustness}
  \label{tab:overview-quant}
\end{table*}

Gorsline et al. investigated the effect of weight quantization on robustness in~\cite{gorsline2021adversarial}. 
They experimented on MNIST~\cite{lecun1998gradient} and a two-spiral classification problem.% Besides early stopping, they did not use regularization methods or adversarial defenses. 
They concluded with the observation that quantization does not affect robustness if the adversarial attack exceeds a critical strength.

Finally, Varghese et al.~\cite{varghese2022joint} introduced a novel hybrid compression approach that combines pruning and quantization and studied the relationships between robustness and compression.  They investigated the more complex task of semantic segmentation for automated driving. In contrast to the other works, the authors investigated corruption robustness, not adversarial robustness. By corruption, they refer to augmentations caused by real-world events (e.g., noise, blur, or weather conditions). They observed improved robustness of the compressed DeepLabv3+~\cite{chen2018encoder} network compared to the reference network.

In summary, naive quantization without AT has demonstrated both negative ~\cite{lin2019defensive} and positive~\cite{wijayanto2019towards} impact on adversarial robustness. If quantization was combined with AT, a positive effect was observed in several works~\cite{galloway2017,wijayanto2019towards,lin2019defensive}. Moreover, AT was shown to improve quantization itself~\cite{lin2019defensive}.

\section{Relationship between Pruning and Robustness} \label{pruning}
An overview of the works that focus on pruning and robustness is given in Table \ref{tab:overview}. 
We divide the considered approaches into three groups: (1) works that examine the intrinsic relationships between pruning and robustness, (2) works proposing novel approaches via a combination of static pruning with robust training, and (3) the dynamic pruning approach, incorporating adversarial robustness as a training objective. 
%Further related topics discussed in Section \ref{sub:ticket} include the trade-off between compression, accuracy, and robustness, comparing pruned models with models of comparable size trained from scratch, and the lottery ticket hypothesis. \\

\begin{table*} 
  \centering
  \resizebox{1.0\linewidth}{!}{%
 % {\def\arraystretch{1.5}\tabcolsep=5pt
  \begin{tabular}{|l|c | c|c|c|c|c|c|c|c|c|} 
  \hline
   \textbf{Author} & \textbf{Year} & \textbf{Ref} & \textbf{Architectures} & \textbf{Dataset}  &  \textbf{Baseline} & \textbf{Pruning Strategy} &   \makecell{\textbf{Attack}} &   \textbf{Attack Method} & \textbf{AT} \\\hline \hline

   Wang et al. & 2018 &\cite{wang2018adversarial} & \makecell{CNN, \\Wide Resnet-28-4~\cite{zagoruyko2016wideresnet}} & \makecell{MNIST~\cite{lecun1998gradient} \\ CIFAR-10~\cite{krizhevsky2009learning}} & \makecell{Non-pruned model} & \makecell{Magnitude-based \\ weight and filter pruning}  & \makecell{White-box \\ Black-box} & \makecell{FGSM~\cite{explaining2014}, PGD~\cite{madry2017towards}, \\ Papernot's attack~\cite{papernot2016limitations}} & \makecell{$\checkmark$ $\times$}  \\ \hline
%      %In \cite{guo2018sparse} the authors demonstrate that there exists intrinsic relationships between  the sparsity of DNN-based classifcation models and their adversarial robustness.
%      %The authors demonstrate, that an appropriately higher model sparsity implies better robustness of nonlinear DNNs, whereas over-sparsified models can be more difficult to resist adversarial examples.
    Guo et al. & 2018 & \cite{guo2018sparse} &   \makecell{LeNet-300-100~\cite{lecun1998gradient},\\ LeNet-5~\cite{lecun1998gradient} \\  ResNet-32~\cite{he2016deep} \\
    VGG-like ResNet~\cite{neklyudov2017structured} } & \makecell{MNIST~\cite{lecun1998gradient} \\CIFAR-10~\cite{krizhevsky2009learning}}&\makecell{Dense models} & \makecell{Progressive pruning} & \makecell{White-box} & \makecell{FGSM, rFGSM~\cite{explaining2014} \\DeepFool~\cite{deepfool2016} \\C\&W $L_{2}$~\cite{carlini2017towards}} &\makecell{$\times$} \\ \hline

    Jordao et al. & 2021 & \cite{jordao2021effect}
    %In \cite{jordao2021effect} the authors empirically show that pruning filters and/or layers of convolutional networks increase their adversarial robustness, while preserving their generalization; thus, it efficiently  satisfies the dilemma between robustness and generalization.
    & \makecell{ResNet56~\cite{he2016deep}\\MobileNetV2~\cite{sandler2018mobilenetv2}, \\ VGG16~\cite{simonyan2014very} } & \makecell{ImageNet-C~\cite{russakovsky2015imagenet} \\CIFAR10~\cite{krizhevsky2009learning}} &\makecell{Other defense mechanisms\\ (style transfer, MixUp~\cite{mixup} \\Cutout~\cite{devries2017improved}, CutMix~\cite{yun2019cutmix} \\Shape-Texture) } &\makecell{Pruning with different\\ criteria ($\ell_{1}$-norm, \\expectedABS~\cite{min2020dropnet}, HRank~\cite{lin2020hrank},\\ KL-divergence~\cite{luo2020neural},\\ partial least squares~\cite{jordao2020deep} )}&White-box&\makecell{FGSM~\cite{explaining2014} \\semantic-preserving \\ transformations~\cite{hendrycks2019benchmarking}, \\simple occlusions, \\transfer attacks}  &\makecell{$\times$} \\ \hline \hline
    
%% Second group
    Liao et al. & 2022 & \cite{liao2022achieving} &
    % Architectures
    \makecell{
    VGG16~\cite{simonyan2014very}, \\
    ResNet18~\cite{he2016deep}, \\
    DenseNet-BC~\cite{huang2017densely} \\
    DenseNet121~\cite{huang2017densely} \\
    }
    &
    % Dataset
    \makecell{
    CIFAR-10~\cite{krizhevsky2009learning} \\
    CIFAR-100~\cite{krizhevsky2009learning} \\
    Tiny-ImageNet~\cite{russakovsky2015imagenet}
    }
    &
    % Baseline
    \makecell{
    Non-pruned models with AT \\ with SOTA clean and  \\ adversarial accuracy
    }
    &
    % Pruning
    \makecell{
    Global unstructured pruning, \\
    local unstructured pruning, \\
    filter pruning,  \\
    network slimming
    }
    &
    % Attack Type
    \makecell{
    White-box
    }
    &
    % Attack Method
    \makecell{
    $L_{\infty}$-PGD~\cite{madry2017towards}
    }
    &
    % AT
    \makecell{
    $\checkmark$
    } \\ \hline
% %%%%%%%%%%%%%%%%%%%%%%%%%%%%%%%%%%%%%%%%%%%%%%%%%%%%%%%%%%%%%%%%%%%%%%%%%%%%%%%%%%%%%%%%%%%%%%%
% %%%%%%%%%%%%%%%%%%%%%%%%%%%%%%%%%%%%%%%%%%%%%%%%%%%%%%%%%%%%%%%%%%%%%%%%%%%%%%%%%%%%%%%%%%%%%%%
% %%%%%%%%%%%%%%%%%%%%%%%%%%%%%%%%%%%%%%%%%%%%%%%%%%%%%%%%%%%%%%%%%%%%%%%%%%%%%%%%%%%%%%%%%%%%%%%
%     \\[.5\normalbaselineskip]
%     %\hline
%     \midrule
%     % Ref
%     \parbox[t]{0.3cm}{
    Gui et al. & 2019 & \cite{gui2019model}
    &
    % Architectures
    \makecell{
    LeNet~\cite{lecun1998gradient},\\
    ResNet34~\cite{he2016deep}, \\
    Wide ResNet~\cite{zagoruyko2016wideresnet} % Sergey Zagoruyko and Nikos Komodakis. Wide residual networks. arXiv preprint arXiv:1605.07146, 2016.
    }
    &
    % Dataset
    \makecell{
    MNIST~\cite{lecun1998gradient} \\
    CIFAR-10~\cite{krizhevsky2009learning} \\
    CIFAR-100~\cite{krizhevsky2009learning} \\
    SVHN~\cite{netzer2011reading} \\
    }
    &
    % Baseline
    \makecell{
    Compressed models,  \\ with and without AT,\\models with AT
    }
    &
    % Pruning
    \makecell{
    Magnitude-based
    }
    &
    % Attack Type
    \makecell{
    White-box
    }
    &
    % Attack Method
    \makecell{
    FGSM~\cite{explaining2014}, PGD~\cite{madry2017towards},     \\
    WRM~\cite{sinha2017certifying}
    }
    &
    % AT
    \makecell{
    $\checkmark$ $\times$
    } \\ \hline

    Ye et al. & 2019 & \cite{ye2019adversarial} & \makecell{LeNet~\cite{lecun1998gradient},\\VGG-16~\cite{simonyan2014very}, \\ResNet-18~\cite{he2016deep}} & \makecell{MNIST~\cite{lecun1998gradient} \\CIFAR-10~\cite{krizhevsky2009learning}} & \makecell{Non-pruned models \\ with and without AT} & \makecell{ ADMM~\cite{han2015learning} with filter,  \\ column, irregular P}
    &
    % Attack Type
    \makecell{
    White-box
    }
    &
    % Attack Method
    \makecell{
    PGD~\cite{madry2017towards} \\ 
    C\&W $L_{\infty}$~\cite{carlini2017towards} \\ 
    Transfer attacks
    }
    &
    % AT
    \makecell{
    $\checkmark$ $\times$
    } \\ \hline

    Sehwag et al. & 2020 & \cite{sehwag2020hydra} & \makecell{VGG-16~\cite{simonyan2014very}, \\Wide-ResNet-28-4~\cite{zagoruyko2016wideresnet},\\CNN-small, CNN-large} & \makecell{CIFAR-10~\cite{krizhevsky2009learning} \\SVHN~\cite{netzer2011reading} \\ImageNet~\cite{russakovsky2015imagenet}} & \makecell{Models with AT,\\ADMM~\cite{han2015learning}-pruned models} & \makecell{HYDRA} & \makecell{White-box} & \makecell{PGD~\cite{madry2017towards}} & \makecell{$\checkmark$} \\ \hline \hline
    
%% Third group
    Hu et al. & 2020 & \cite{hu2020triple} &\makecell{ SmallCNN \\ResNet-38~\cite{he2016deep} \\ MobileNet-V2~\cite{sandler2018mobilenetv2}} & \makecell{MNIST~\cite{lecun1998gradient} \\CIFAR-10~\cite{krizhevsky2009learning}} & \makecell{Non-pruned models with AT,\\ SSS-pruned\cite{huang2018data} models \\with AT,  ATMC~\cite{gui2019model}} & \makecell{Dynamic pruning\\ with RDI-Nets, SSS~\cite{huang2018data}} & \makecell{White-box} & \makecell{PGD~\cite{madry2017towards} \\FGSM~\cite{explaining2014} \\WRM~\cite{sinha2017certifying}} & \makecell{$\checkmark$} \\\hline
  \end{tabular}
  }
  \caption{Overview of the publications analyzing the relationship between \textbf{pruning} and adversarial robustness}
  \label{tab:overview}
\end{table*}

% Kurz auf Marzi et al eingehen? (binary classification, SVM)
% Kurz auf Mantowsky eingehen
\subsection{Effects of Pruning on Robustness}
The first group of works aims at studying the general effects of pruning on adversarial robustness. In the theoretical and empirical analyses, particular attention was paid to the question of whether pruning offers inherent protection against adversarial attacks. 

% Kurz die Gruppe beschreiben
%
%
%
% Erste Untersuchung, mit und ohne AT, wenig Architekturen, nicht unbedingt state-of-the art pruning etc.
Wang et al.~\cite{wang2018adversarial} conducted the first analysis regarding the adversarial robustness of pruned deep neural networks. The work was not published because the experimental evidence was not grounded enough. 
The effects of pruning on robustness and the impact of AT on pruned networks were investigated. 
Naturally trained models were compared to their original networks. The accuracy of a pruned model was similar to the accuracy of an original network. The robustness of a pruned network under FGSM and Papernot's attacks was worse than the robustness of an original network. Neither the pruned nor the original model could withstand the PGD attack. The authors suspected, that pruning reduces the network capacity, which in turn reduces its robustness. 
Then, the authors performed AT with FGSM and PGD along with the network pruning procedure and compared these models to their respective adversarially trained original networks.  
They observed that highly pruned networks can become considerably robust, while weight pruning allows more compression than filter pruning, and PGD leads to more robust models than FGSM. 

In additional experiments with a Wide ResNet~\cite{zagoruyko2016wideresnet} on CIFAR-10~\cite{krizhevsky2009learning}, the authors observed an interesting result. The PGD-trained network that was moderately pruned (less than 50\% of the parameters) was slightly more accurate and more robust than the respective original network. The robustness of the highly pruned network (80\% to 94\% of the weights) was higher than the original, but the accuracy on natural images dropped simultaneously. 
With an increasing compression rate, the robustness of the model drops earlier than the classification accuracy. The authors observed that with the training procedures applied, a model cannot be both highly robust and pruned simultaneously. 

Another early work that studied the intrinsic relationships between the sparsity achieved through weight and activation pruning and the adversarial robustness of DNNs is by Guo et al.~\cite{guo2018sparse}. Their analysis is one of the few works that examine the effects of pure pruning without AT on adversarial robustness. 
The authors trained different architectures and evaluated their robustness under various $l_{2}$ and $l_{\infty}$ white-box attacks. For the evaluation of the robustness of the models, the authors suggested two metrics that describe the ability to resist $l_{2}$ and $l_{\infty}$ attacks, respectively.
First, they pruned the weights of the dense reference networks and compared the robustness of the pruned networks to the original ones. Sparse DNNs are prone to be more robust against $l_{\infty}$ (FGSM and rFGSM~\cite{explaining2014}) and $l_{2}$ (DeepFool~\cite{deepfool2016}, C\&W $L_{2}$~\cite{carlini2017towards}) attacks until the sparsity reaches some thresholds, above which the capacity of the pruned models degrades. This observation is consistent with the observations from~\cite{wang2018adversarial} described above. The authors verified their results additionally with the attack-agnostic CLEVER~\cite{weng2018evaluating} scores. %At this point, it should be noted that the authors did not use the PGD attack in their experiments.
%Then they examined the relationship between robustness and activation sparsity. 
They observed positive correlations between activation sparsity in a certain range and robustness. 
The authors suggested taking care and avoiding sparsity rates that are too high and concluded that sparse nonlinear DNNs can be more robust than their dense counterparts if the sparsity is within a certain range. 

Similar to the work by Guo et al.~\cite{guo2018sparse}, Jordao and Pedrini~\cite{jordao2021effect} studied the intrinsic effect of pruning on the adversarial robustness of deep convolutional networks without AT.
However, unlike~\cite{wang2018adversarial, guo2018sparse}, the authors did not examine the trade-off between robustness, accuracy, and compression but the relationship between generalization and robustness. They observed that pruning preserves generalization. 
The authors pruned filters and layers from several reference architectures based on different pruning criteria.
After pruning, they fine-tuned the compressed networks with augmented data. 
First, they compared the accuracy and robustness of the dense reference networks to their pruned counterparts (filters, layers, and both) under different attacks. 
Overall, they observed that pruning improves robustness without sacrificing generalization. Similar to~\cite{guo2018sparse}, the authors did not use the PGD attack in their experiments. 

Furthermore, they could not observe a superior pruning strategy with respect to all attacks. 
Then, they demonstrated that removing single filters can improve the robustness without adjusting the network parameters. 
%The authors also investigated whether training from scratch is better regarding adversarial robustness than pruning and fine-tuning. 
They also observed that fine-tuning leads to increased adversarial robustness than training from scratch. 
%Finally, they compared pruned networks to other defense mechanisms. % like Stylized, MixUp, Cutout, CurMix, and Sharpe-Texture. 
When comparing the pruned network to other defense mechanisms, they observed that pruning obtained one of the best average improvements. They suggested combining pruning with other defense mechanisms to achieve more robust and efficient networks. 
The authors concluded that pruning filters or layers (or both) increase the adversarial robustness of convolutional networks.

In summary, both negative~\cite{wang2018adversarial,gui2019model} and positive~\cite{guo2018sparse,jordao2021effect} effect of pruning on robustness were seen in the experiments, although studies leading to the latter provided significantly more empirical evidence. Both papers observing positive effects~\cite{guo2018sparse,jordao2021effect} have used retraining -- this confirms again that omitted retraining strongly weakens robustness. On the other hand, these works did not provide results for the PGD, making comparing the pieces of evidence difficult.

\subsection{Combined Compression-Robustness Methods}
Various combined compression-robustness approaches were proposed, with network pruning performed before, after, or alternately with AT.
% Gruppe beschreiben + erklären dass aus den Experimenten auch teilweise Rückschlüsse auf die allgemeine Beziehung Pruning - Robustness gezogen werden können
Liao et al.~\cite{liao2022achieving} theoretically proved the correlation between weight sparsity and adversarial robustness and showed in experiments that weight sparsity improves robustness with AT. %Additionally, they proposed a novel adversarial training method. 
%They adversarially trained reference networks with different architectures and pruned each network with two structured and two unstructured pruning methods. Then they compared the dense reference networks to their pruned counterparts. 
They showed that pruning does not affect the model robustness negatively in some adversarial settings. Furthermore, they demonstrated, that the robustness can be improved with AT after pruning. Overall, the proposed novel AT method that includes pruning was shown to lead to sparse networks with better performance than their dense counterparts. 

In~\cite{gui2019model} the authors stated, that they describe the first framework that connects model compression with adversarial robustness. They proposed their Adversarially Trained Model Compression (ATMC) framework, which includes pruning, quantization, and AT. 
ATMC was compared to adversarially trained, pruned, adversarially trained, and pruned, as well as adversarially trained, pruned, and adversarially retrained models.
Their results support the existence of a trilateral trade-off between robustness, accuracy, and compression.
Analogously to~\cite{wang2018adversarial, guo2018sparse}, the authors concluded, that if robustness is taken into account, model compression can maintain accuracy and robustness, whereas naive model compression may decrease adversarial robustness. 

A similar approach is proposed by Ye et al~\cite{ye2019adversarial}. The authors proposed a framework of concurrent AT and weight pruning. 
To compare weight pruning and training from scratch, they adversarially trained models of different architectures with various scaling factors. %. (1, 2, 4, 8, 16). 
Then, the authors pruned the filters of each network with the proposed framework. Each reference network was pruned to the respective smaller scaling factors. The authors summarized that pruned networks can have high accuracy and robustness, which can be lost if a network with a comparable size is adversarially trained from scratch. 
Framework evaluation under different pruning schemes and transfer attacks has demonstrated, that irregular pruning performs the best and filter pruning performs the worst. Interestingly, the pruned model turned out to be more robust to transfer attacks than the respective dense network. 
%Furthermore, they suggested that models are most vulnerable to adversarial examples generated by themselves. 

In~\cite{sehwag2020hydra}, pruning is formulated as an empirical risk minimization problem, while the minimization problem can be integrated with various robust training objectives like AT. The authors demonstrated that pruning after training helps to achieve state-of-the-art accuracy and robustness. %Furthermore, they established a strong baseline with the adversarial training approach from Carmon et al.~\cite{carmon2019unlabeled}. 
The proposed method (HYDRA) incorporates the AT approach by Carmon et al.~\cite{carmon2019unlabeled}, although other robust training objectives are possible. 
The authors observed improved compression, accuracy, and robustness compared to the baseline networks and previous work like the ADMM~\cite{han2015learning}-based approach by Ye et al.~\cite{ye2019adversarial}.
The authors advocated for formulating pruning as an optimization problem that integrates the robust training objective. They identified the performance gap between non-pruned and pruned networks as an open challenge. 

In summary, two works~\cite{ye2019adversarial,sehwag2020hydra} observed a significantly higher robustness of pruned networks compared to compact networks of comparable size. Furthermore, the authors concluded that pruned networks can, after all, exhibit similar robustness to their dense reference networks. 

Furthermore, the results overall indicate that the effect of pruning on robustness varies in magnitude depending on whether we are comparing networks of the same capacity or networks of different capacities. Retraining the pruned models seems to be a crucial factor in that view. It was observed that most networks show a higher robustness when retrained after pruning, compared to the networks for which no retraining was performed. %Interestingly, some networks did not benefit from retraining.  - 	but I don't have an answer to this question either.

\subsection{Dynamic Pruning and Robustness}
% Kurz hier erklären

Hu et al.~\cite{hu2020triple} proposed the first dynamic approach to improve network efficiency, accuracy, and robustness and called it Robust Dynamic Inference Networks (RDI Nets). These networks are based on the work of Kaya et al.~\cite{kaya2019shallow}. RDI-nets stop inference in early layers.
In their experiments, the authors evaluated three adversarially (PGD) trained models against their respective RDI nets using three white-box attack algorithms, which were executed in three proposed attack forms.  
Then the authors compared the RDI-nets to defended sparse networks, i.e., networks that were compressed with a state-of-the-art network pruning method Sparse Structure Selection (SSS)~\cite{huang2018data} and then adversarially retrained (PGD). Furthermore, they compared their RDI nets to the latest ATMC algorithm~\cite{gui2019model}.
The pruning + defense baseline has demonstrated superior robustness compared to the respective dense reference network.
The authors concluded with the statement that they achieved better accuracy, stronger robustness, and computational savings of up to 30\%. It should be noted, however, that dynamic pruning does not reduce the model size, but can only achieve efficiency gains in terms of the required computing resources.

\subsection{Connection to the Lottery Ticket Hypothesis} 
\label{sub:ticket}
%Different hypotheses about general pruning exist. 
The lottery ticket hypothesis by Frankle et al.~\cite{frankle2018lottery} states that randomly initialized networks contain subnetworks ("the winning tickets"). When trained in isolation, these subnetworks can reach test accuracies comparable to the reference network in a less or equal number of iterations. The initial weights of these winning tickets make training particularly effective. The only meaning of weight pruning is thus the effective initialization of the final pruned model. 

In contrast, Liu et al.~\cite{liu2018rethinking} observed that the winning ticket initialization does not bring improvement over random initialization. They showed that training from scratch gave comparable or better performance than SOTA pruning algorithms, thus making the original network's inherited weights useless. The meaning of weight pruning is thus the pruned architecture itself. They suggested that pruning can be a useful architecture search paradigm, but the pruned network should be trained with random initialized values. 

A few works examined these hypotheses with respect to adversarial robustness. In particular, Ye et al.~\cite{ye2019adversarial} observed that training from scratch cannot achieve robustness and accuracy simultaneously, even with inherited initialization, which contradicts the lottery ticket hypothesis. In contrast, Liao et al.~\cite{liao2022achieving} concluded that preferable adversarial robustness can be achieved through the lottery ticket settings. They argue that they search for the winning ticket by iterative global unstructured pruning, while Ye et al.~\cite{ye2019adversarial} used filter pruning. Jordao et al.~\cite{jordao2021effect} showed that fine-tuning leads to better robustness than the winning ticket.

%network did not provide positive average improvements while fine-tuning achieved the best gains.  They concluded that fine-tuning often leads to better robustness.
%The authors trained a pruned network with randomized initialization for the same number of epochs as its unpruned counterpart and trained a second pruned network by doubling the epochs. Furthermore, they trained a winning ticket network with inherited weights from the unpruned network and trained a pruned network for a few epochs with the current parameters of the pruned network (fine-tuning).Then the authors examined the robustness of these networks against different attacks. 

Finally, Sehwag et al.~\cite{sehwag2020hydra}  demonstrated the existence of hidden sub-networks that are more robust than the original network. They showed that highly robust sub-networks exist even within non-robust networks.

 \section{Conclusion}
In this work, we reviewed and compared the existing works exploring the relationship between model compression methods (quantization and pruning) and adversarial robustness. 

Throughout all experiments, it was shown that naive pruning and quantization can reduce robustness. Furthermore, as long as networks are compressed within certain limits, pruning may preserve or even improve robustness, especially when comparing compressed and compact models of the same size. 

Moreover, the reviewed works showed that combining model compression and robustness in AT is possible. However, a trade-off exists between compression ratio, accuracy, and robustness. It was observed relatively consistently that once a critical compression ratio is exceeded, first the robustness and then the accuracy decrease. Some authors explain that robustness thus requires a greater capacity than accuracy.

Overall, many reviewed works agree that compression must be performed carefully. Simple, straightforward compression can also have negative effects on robustness; some authors, therefore, also suggest that robustness should be taken into account in the evaluation of new compression methods.

\section*{Acknowledgment}

This research is funded by the German Federal Ministry of Education and Research within the project "GreenEdge-FuE“, funding no. 16ME0517K.
%\footnote{\url{https://www.elektronikforschung.de/projekte/greenedge-fue}}.

\bibliographystyle{IEEEtran}
{\tiny \bibliography{references.bib}}

\end{document}